\documentclass[runningheads]{llncs}

\usepackage{eccv}

\usepackage{eccvabbrv}

\usepackage{graphicx}
\usepackage{booktabs}
\usepackage{pifont}
\usepackage{makecell}
\usepackage{multirow}
\usepackage{arydshln}
\usepackage[accsupp]{axessibility}  

\usepackage{hyperref}

\usepackage{orcidlink}
\begin{document}

\title{Interleaving One-Class and Weakly-Supervised Models with Adaptive Thresholding for Unsupervised Video Anomaly Detection} 

\titlerunning{Interleaving OCC and WS Models with Adaptive Thresholding for UVAD}

\author{Yongwei Nie\inst{1}\orcidlink{0000-0002-8922-3205} \and
Hao Huang\inst{1}\orcidlink{0009-0004-6363-1229} \and
Chengjiang Long\inst{2}\orcidlink{0000-0003-1584-7290} \and
Qing Zhang\inst{3}\orcidlink{0000-0001-5312-2800} \and
Pradipta Maji\inst{4}\orcidlink{0000-0002-8288-8917} \and
Hongmin Cai\inst{1}\inst{\thanks{Hongmin Cai is the corresponding author. Email: hmcai@scut.edu.cn}}\orcidlink{0000-0002-2747-7234}}

\authorrunning{Y.~Nie et al.}

\institute{South China University of Technology, China \and
Meta Reality Labs, USA \and
Sun Yat-sen University, China \and
Indian Statistical Institute, India
}

\maketitle

\begin{abstract}
  Video Anomaly Detection (VAD) has been extensively studied under the settings of One-Class Classification (OCC) and Weakly-Supervised learning (WS), which however both require laborious human-annotated normal/abnormal labels. In this paper, we study Unsupervised VAD (UVAD) that does not depend on any label by combining OCC and WS into a unified training framework. Specifically, we extend OCC to weighted OCC (wOCC) and propose a wOCC-WS interleaving training module, where the two models automatically generate pseudo-labels for each other. We face two challenges to make the combination effective: (1) Models' performance fluctuates occasionally during the training process due to the inevitable randomness of the pseudo labels. (2) Thresholds are needed to divide pseudo labels, making the training depend on the accuracy of user intervention. For the first problem, we propose to use wOCC requiring soft labels instead of OCC trained with hard zero/one labels, as soft labels exhibit high consistency throughout different training cycles while hard labels are prone to sudden changes. For the second problem, we repeat the interleaving training module multiple times, during which we propose an adaptive thresholding strategy that can progressively refine a rough threshold to a relatively optimal threshold, which reduces the influence of user interaction. A benefit of employing OCC and WS methods to compose a UVAD method is that we can incorporate the most recent OCC or WS model into our framework. Experiments demonstrate the effectiveness of the proposed UVAD framework. Our code is available at \href{https://github.com/benedictstar/Joint-VAD}{https://github.com/benedictstar/Joint-VAD}.
\end{abstract}

\section{Introduction}
\label{sec:intro}

Video Anomaly Detection (VAD) is a task that identifies abnormal events in a video, where the abnormal event could be a fire alarm, a flaw in an industrial product, or a traffic accident, etc.
Most previous VAD methods fall in two categories: One-Class Classification (OCC) methods~\cite{liu2021hybrid,liu2018future,gong2019memorizing,Sun_2023_CVPR,wang2022video,hirschorn2023normalizing} and Weakly-Supervised (WS) approaches~\cite{sultani2018real,tian2021weakly,wan2020weakly,wu2022self,Lv_2023_CVPR,Zhang_2023_CVPR}. Both kinds of approaches require human-annotated labels for training \footnote[1]{OCC models are trained on normal data only, which requires human labor to exclude abnormal events in preparing the training dataset.}. Supervised VAD methods have two prominent problems. 
First, abnormal events are sparse and difficult to recognize, making the annotation of the training dataset a highly labor-intensive task.  
Secondly, the categories of abnormal events are unbounded, while human can only collect a limited scope of them, yielding the risk that the trained VAD methods cannot handle unobserved abnormal events. 

Considering the above problems, this paper studies Unsupervised VAD (UVAD) that does not rely on labels. Despite the importance of UVAD, few methods have been developed except \cite{zaheer2022generative} and its subsequent works \cite{al2024coarse,tur2023exploring} to the best of our knowledge. To achieve unsupervised learning, the method of \cite{zaheer2022generative} adversarially trains a generator and a discriminator, where the generator is an autoencoder-based reconstruction model and the discriminator is a fully-connected (FC) binary classifier. While the proposed network architecture is sophisticated, the adopted autoencoder and FC networks are relatively weak which limit the performance of the approach.  

In this paper, we propose a new UVAD framework. Our key idea is that the techniques of OCC and WS approaches are rapidly developed, while a UVAD method is usually implemented by training two VAD models alternately generating pseudo labels for each other. This motivates us to propose a UVAD framework that directly interleaves the training of a pair of OCC and WS models. This allows us to incorporate the recent advances in the two hot research fields to implement a UVAD method. We have also attempted to combine two OCC or two WS models, but found two homogeneous models are less effective. 

While combining OCC and WS methods, we encounter two problems preventing it from effective. First, we observe the OCC or WS model's performance fluctuates occasionally during the training process, which affects the final accuracy they can achieve. This is partly because the pseudo labels generated by each other are changed frequently, making the training not stable. To solve this problem, we extend OCC to weighted OCC (wOCC), and propose a wOCC-WS interleaving training module. Previous OCC model is trained on normal data, while our wOCC model is trained with soft labels in the range of $[0,1]$. The soft labels are more consistent across adjacent training cycles, making the performance of wOCC more stable than OCC under changing pseudo labels, thus suppressing the fluctuation effect. 
Second, we still have to set a threshold to partition normal and abnormal pseudo labels for the WS model, making our method susceptible to a user-provided hyperparameter. 
To reduce the influence of user interaction, we repeat the interleaving training module multiple times, during which we propose an adaptive threshold mechanism that finds an optimal threshold given just a rough initial value from the user. 

To conclude, the contributions of this work include:
\begin{itemize}
    \item We propose a new UVAD framework that interleaves the training of a pair of OCC and WS models. 
    \item We improve OCC to wOCC which stabilizes the training of the UVAD framework, and also propose an adaptive thresholding strategy to alleviate the influence of user interaction. 
    \item Our UVAD framework is flexible to employ different pairs of  OCC and WS models. Experiments on three OCC and two WS models demonstrate the effectiveness of our method. 
\end{itemize}

\section{Related Work}
\textbf{Unsupervised VAD (UVAD).} Unsupervised VAD requires identifying anomalies from data containing both normal and abnormal samples without any annotation. This is a challenging new task that is almost unexplored in the literature. Although significant progress has been made in OCC and WS VAD, directly applying them independently to address UVAD does not yield good results. Zaheer et al.~\cite{zaheer2022generative} firstly raise this task and propose to solve UVAD by generative cooperative learning (GCL). Differently, our proposed method modifies both the OCC and WS VAD models and combines the modified two models together to learn from unmarked training data.

\textbf{One-Class Classification VAD (OCC).} OCC-based VAD approaches only have access to normal data and try to model normal data to identify behaviors that are significantly different from normal behaviors as anomalies. Early works use hand-crafted features to help detect anomalies~\cite{basharat2008learning,medioni2001event,wang2014learning,zhang2009learning}. With the rapid development of deep learning, recent methods turn to normal representations extracted by using a deep neural network~\cite{tudor2017unmasking,pang2020self,smeureanu2017deep,wang2019gods}. Some methods identify normal patterns by using the reconstruction/prediction model to reconstruct the representations~\cite{liu2021hybrid,liu2018future,gong2019memorizing,zhao2017spatio,burlina2019s,ionescu2019object,morais2019learning,nguyen2019anomaly,park2020learning,sohrab2018subspace,Yang_2023_CVPR,Sun_2023_CVPR,Yan_2023_ICCV}. 
These models may lead to well-reconstructed anomalies, thereby limiting the performance of the OCC-based methods. Other OCC-based methods turn to identify normal representations by using proxy tasks~\cite{georgescu2021background,barbalau2023ssmtl++,wang2022video,lorre2020temporal,shi2023video}. What's more, recently, Hirschorn and Avidan~\cite{hirschorn2023normalizing} propose to build a multivariate Gaussian distribution for normal data and detect instances deviating from this distribution as anomalies. However, the above methods treat all samples equally, ignoring the potential differences in importance between normal samples.

\textbf{Weakly-Supervised VAD (WS).} In weakly-supervised VAD, video-level annotations are available in the training stage. Sultani et al~\cite{sultani2018real} first propose to use the video-level labels and solve WS VAD based on multi-instance learning (MIL) framework~\cite{andrews2002support,dietterich1997solving}. Since then, many works~\cite{zhang2019temporal,wan2020weakly,wu2022self,Lv_2023_CVPR,Zhang_2023_CVPR,sapkota2022bayesian} have viewed the WS VAD task as a MIL problem. Tian et al.~\cite{tian2021weakly} develop to extend MIL to Top-$k$ MIL method by training with a robust temporal feature magnitude (RTFM) loss function. The methods mentioned above directly train in the supervision of video-level labels and the coarse-grained supervision limits the accuracy of WS models. Recently, two-stage self-training methods~\cite{feng2021mist,li2022self,zhong2019graph,wu2020not} adopt a two-stage pipeline to use more fine-grained labels to supervise the networks more strictly. The WS model in our framework shares a similar idea with \cite{feng2021mist} and is tailored to use finer-grained snippet-level labels.

\section{Method}
\begin{figure}[t]
    \centering
    \includegraphics[width=1.0\textwidth]{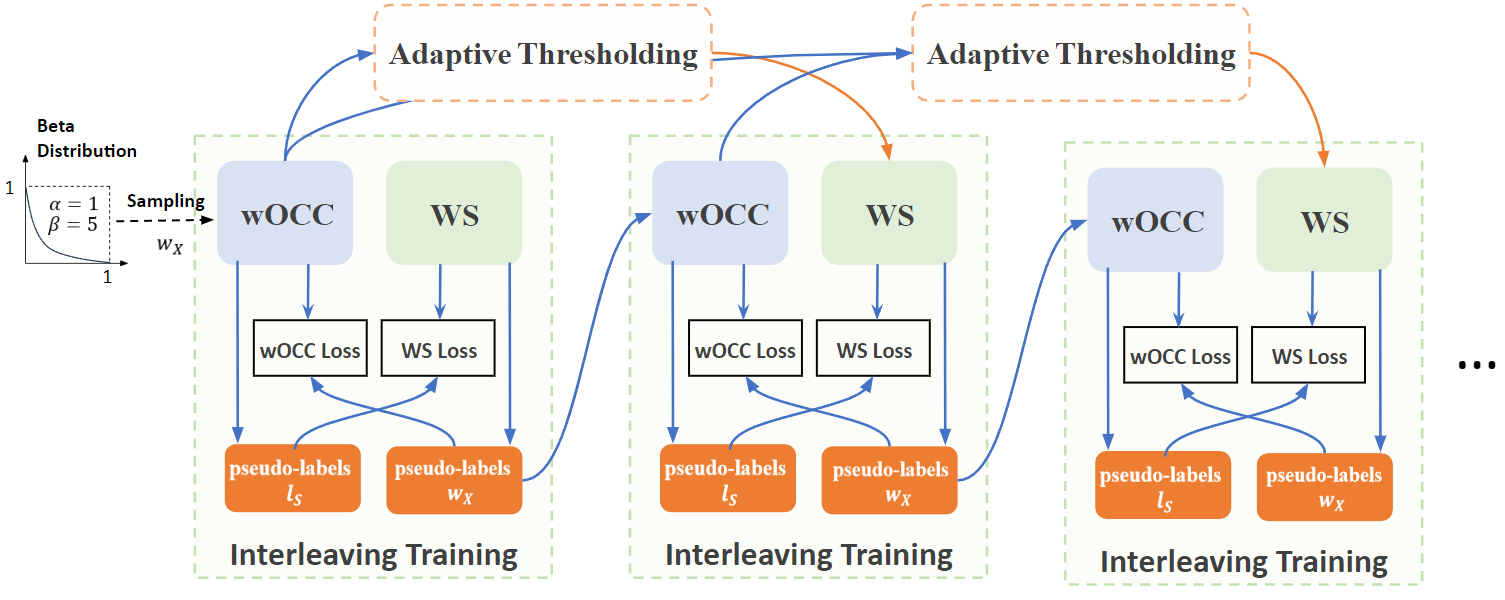}
    \caption{Pipeline of our proposed interleaved framework. The wOCC-WS interleaving training module comprises a weighted OCC (wOCC) model and a WS model, using the output of each model as the pseudo-labels to supervise the other model. Further, the training module is repeated multiple times with the adaptive threshold until the stopping criterion we set. To start the first training module, we randomly sample $w_X$ to train the wOCC model and assign a large value as the threshold for the WS model. Once a training module converges, the last WS model of it supplies $w_X$ to start the next training module and adaptive thresholding takes the outputs of all previous wOCC models into consideration for updating the threshold. } 
    \label{fig:overview}
\end{figure}
\subsection{Overview}
Our proposed UVAD framework interleaves the training of a pair of OCC and WS models. In theory, any OCC or WS methods can be incorporated into our framework. Without loss of generality, we take STG-NF~\cite{hirschorn2023normalizing} (a recent OCC model) and RTFM~\cite{tian2021weakly} (a recent WS model) as examples to introduce our method in the following sections. We test many other OCC/WS models in 
Section~\ref{sec:discussion}. 

Figure~\ref{fig:overview} shows an overview of our framework. The basic buildingblock of our method is the wOCC-WS interleaving training module. In the module, the outputs of each model are used as pseudo labels for defining the loss function training the other model. After one module converges, we re-initialize another module to run the interleaving training again with a difference that the threshold of the WS model is updated by an adaptive threshold adjustment mechanism based on wOCC models trained in previous modules. Since wOCC has more stable behaviors than WS, we always initialize the wOCC model to start the training of a module. For the first module, we use a random initialization based on Beta distribution sampling. For other modules, we use the results of WS of the preceding module for the initialization.

\subsection{wOCC-WS Interleaving Training Module}
\label{sec: wOCC-WS Interleaving Training Module} 
We first introduce the wOCC-WS interleaving training module which comprises a wOCC model and a WS model. 
At the very beginning, we initialize the wOCC model to start the interleaving training. Then, pseudo labels created using wOCC are used to train the WS model. Next, pseudo labels created by the WS model are used to improve the wOCC model. The whole training does not use human annotations. 

\textbf{wOCC Model (STG-NF).} Typically, OCC models are trained only on normal data. Directly using OCC in our UVAD framework may make the training fluctuate, i.e., the performance of the trained models sometimes gets better, but sometimes worse, finally converging to an inferior point. We use 0 to indicate normal samples, and 1 abnormal samples, where 0, 1 are hard binary labels. During training, we need to identify 0-labelled samples to train the OCC model. The hard partitioned labels may be suddenly changed from 0 to 1 or from 1 to 0, yielding the frequent change of the normal samples used to train the OCC model which is the source of the unstable training. To this end, we propose wOCC which relies on soft labels instead of hard binary labels. Soft labels are in the range of $[0,1]$. For example, the soft label 0.7 may be changed to 0.6 but not 0 suddenly. This consistency reduces the fluctuation of the interleaving training.

Taking STG-NF~\cite{hirschorn2023normalizing} (an OCC model) as an example, we introduce how to improve it to wOCC. STG-NF extracts a set of objects $\mathcal{X}=\{X_1, \cdots, X_i, \cdots\}$ from the training videos, where $X_i$ is a sequence of poses of a person. Each object $X_i$ is the basic building block processed by STG-NF. Originally, STG-NF minimizes the negative log-likelihood across the normal data:
\begin{equation}
  \mathcal{L}_{occ} = -log(p_Z(f_{STG-NF}(X_i^+))),
  \label{eq: occ-distribution}
\end{equation}
where $X_i^+ \in \mathcal{X}^+$ is sampled from normal data, $Z=f_{STG-NF}(X_i^+)$ is the feature extracted from $X_i^+$ by the STG-NF, $p_Z$ is the established distribution of the normal data. We upgrade STG-NF by introducing the soft labels $w_X=\{w_{X_1},\cdots,w_{X_i},\cdots\}$ as the weight to supervise its training on both normal and abnormal data. Formally, for the wOCC model improved from STG-NF, we minimize the following weighted negative log-likelihood across all the data :
\begin{equation}
  \mathcal{L}_{wocc} = -(1-w_{X_i})log(p_Z(f_{STG-NF}(X_i))),
  \label{eq: weighted occ-distribution}
\end{equation}
where $X_i \in \mathcal{X}$ is sampled from the whole training dataset rather than the normal part and $w_{X_i}\in[0,1]$ is the weight of the object $X_i$ indicating the anomaly degree of the object which serves as the soft label. The object with smaller $w_{X_i}$ means more normal which is better captured by the learned distribution of wOCC. 

\textbf{Pseudo Labels $l_\mathcal{S}$ from wOCC.} Now we use the trained wOCC model to generate pseudo labels $l_{S}=\{l_{S_1},\cdots, l_{S_i}, \cdots\}$ for training the WS model. ``Snippet'' is the basic element processed by WS models. We first split the training videos into snippets, obtaining a set of snippets $\mathcal{S}=\{S_1,\cdots, S_i, \cdots\}$. Our aim is to compute the label $l_{S_i}\in\{0,1\}$ for each snippet $S_i$. Let $x_i$ be the anomaly score of the object $X_i$ computed by the trained wOCC model. The anomaly score $\hat{s}_i$ of a video snippet $S_i$ is computed by averaging all the anomaly scores of objects in the snippet. To generate pseudo label $l_{S_i}$, one way is to check whether $\hat{s}_i$ is larger than a threshold ($l_{S_i}=1$) or not ($l_{S_i}=0$). However, as the training goes, the anomaly scores of snippets may vary in a large range, making the absolute threshold invalid.  
Instead, we sort the snippets according to their anomaly scores in descending order, and get the index of the snippet $S_i$ by ${\rm Rank}(\hat{s}_i)$ in the sorted list. 
Finally, the label $l_{S_i}$ of snippet $S_i$ is computed by 
\begin{equation}
    l_{S_i} = \begin{cases}
        1, \quad {\rm if} \quad {\rm Rank}(\hat{s}_i) < T_{ws}, \\
        0, \quad {\rm otherwise},
    \end{cases}
    \label{eq:ws-threshold}
\end{equation}
where $T_{ws}$ is the relative threshold not affected by concrete anomaly scores.

\textbf{WS Model (RTFM).} With snippet-level pseudo annotations $l_{\mathcal{S}}$ (obtained in Eq.~\ref{eq:ws-threshold}), we now define positive and negative bags and use them to train a WS model. Previously, approaches~\cite{sultani2018real,tian2021weakly} usually treat snippets of the same video as a bag based on video-level labels. Differently, thanks to the snippet-level labels provided by the wOCC model, we can compose positive and negative bags more flexibly. In our method, a positive bag $B^+ = \{S_i^+\}_{i=1}^{C}$ is composed of randomly selected $C=16$ abnormal snippets. Similarly, a negative bag $B^- = \{S_j^-\}_{j=1}^{C}$ is composed of randomly selected $C$ normal snippets. Taking RTFM~\cite{tian2021weakly} as an example, the WS method employs multiple instance learning, requiring the average Top-$k$ maximum magnitudes of features in positive bags exceeds that in negative bags:
\begin{equation}
    \begin{split} 
    \mathcal{L}_{ws} = \operatorname{max} (0,m- d_{RTFM}(B^+) + d_{RTFM}(B^-)) + \\
    \mathcal{L}_{BCE}(B^+) + \mathcal{L}_{BCE}(B^-),
    \label{eq: ws-loss-function}
    \end{split}
\end{equation}
where $m$ is the margin, and $d_{RTFM}(B)$ (where $B$ can be $B^+$ or $B^-$) returns the average of the Top-$k$ maximum magnitudes of features. Finally, $\mathcal{L}_{BCE}$ is the binary cross-entropy-based classification loss:
\begin{equation}
      \mathcal{L}_{BCE}(B)=-Y\operatorname{log}(f_{RTFM}(B)) 
      +(1-Y)\operatorname{log}(1-f_{RTFM}(B)),
  \label{eq: BCE loss}
\end{equation}
where $Y$ is the label of the bag $B$ and $f_{RTFM}(B)$ returns the average of the Top-$k$ maximum anomaly scores. Note that $f_{RTFM}(B)$ is different from $d_{RTFM}(B)$, where $f_{RTFM}(B)$ returns the anomaly scores while $d_{RTFM}(B)$ returns the magnitude of the features. 

\textbf{Pseudo Labels $w_{X}$ from WS.} wOCC model needs soft labels $w_{X}$. With the trained WS model, we compute the anomaly score $\hat{x}_i$ of object $X_i$ by simply setting it as the anomaly score of the snippet containing the object $X_i$. Then, the soft label $w_{X_i}$ is directly set as:
\begin{equation}
    w_{X_i} = \hat{x}_i.
    \label{eq:compute_weights}
\end{equation}
$\hat{x}_i$ is already in the range of $[0,1]$, so we do not need to normalize $w_{X_i}$. 

\textbf{Initializing the Interleaving Training Module}. As discussed above, in our proposed interleaving training module, wOCC training relies on pseudo labels generated by the WS model, and conversely, WS training requires pseudo labels derived from the wOCC. This egg and chicken problem needs to be solved at the very beginning. Our solution is to provide the wOCC model with randomly generated soft labels to start the interleaving training. Usually, the training dataset contains much more normal samples than abnormal ones, which means most soft labels are around 0 while a small portion of them are near 1. Therefore, we randomly sample the weight $w_X$ from the Beta distribution, i.e., $w_X \sim {\rm Beta}(\alpha,\beta)$, where $\alpha=1$ and $\beta=5$. The sampled weights $w_X$ are mostly around 0, and a small portion of them is close to 1, satisfying the prior about the distribution of normal and abnormal data.

\textbf{Convergence Analysis.} We have discussed the mechanism and the way to start the interleaving training module above. Then, in the following, we empirically analyze its convergence, which is validated by the experiment in Figure~\ref{fig:training-loss-curve}. At the very beginning, we randomly initialize the weight $w_X$ to train our wOCC model. Although the random initialization is not precise, the wOCC model can still learn normal patterns from the training data, as we assume there is more normal data than abnormal data in the training dataset. Therefore, the wOCC model can produce relatively reliable pseudo labels for the WS model. This is very important as the WS model has a high requirement on the quality of the pseudo labels. In turn, with the trained WS model, we can provide better pseudo supervision to train a stronger wOCC model than the random initialization. In this way, we achieve mutual improvement between two models until convergence.

The above analysis implies the reason why we do not choose to interleave the training of two OCC models or two WS models. The OCC model can provide reliable initialization, while the WS model explicitly enlarges the gap between normal and abnormal samples. Two OCC models fail to utilize the abnormal data, while for two WS models it is not known how to provide a reliable initialization to start the training.

\subsection{Repeating Procedure with Adaptive Thresholding}
\label{sec:adaptive_thresholding} 
In the above, the wOCC and WS models need to provide pseudo labels to supervise the other. As the wOCC model relies on soft labels $w_{X}$ as defined in Eq.~\ref{eq:compute_weights}, its pseudo labels can be directly provided by the WS model without thresholding. However, as defined in Eq.~\ref{eq:ws-threshold}, we need to specify a threshold for the WS model since it requires binary hard partition labels to construct positive and negative bags. To alleviate the influence of human interaction, we propose a repeating procedure of the interleaving training module which can adaptively refine a user-specified rough threshold to a relatively optimal one.  

\textbf{Starting and running the next interleaving training module.} As shown in Figure~\ref{fig:overview}, we run several interleaving training modules one after another. Recall that we provide $w_X$ by random initialization to start the first interleaving training module as discussed in Section~\ref{sec: wOCC-WS Interleaving Training Module}. To start the next module, we use the final WS model in the last module to provide the labels $w_X$. For the wOCC and WS models in a new interleaving training module, we train them from scratch, not inheriting parameters from the last module. 

\textbf{Adaptive Thresholding.}
The threshold is fixed in each module. Between two modules, we employ an adaptive thresholding mechanism to refine the threshold. As the repeating procedure runs, we obtain a series of thresholds $\{T_{ws}^1,T_{ws}^2,\cdots,T_{ws}^i,\cdots\}$ corresponding to the interleaving training modules. 
The key idea of adaptive thresholding is to ensure:
\begin{equation}
    T_{ws}^1\geq T_{ws}^2\geq\cdots\geq T_{ws}^i\geq \cdots.
    \label{eq:threshold_requirement}
\end{equation}
Starting from a sufficiently large (larger than the optimal threshold) value of $T_{ws}^1$, there must be one threshold $T_{ws}^* \in \{T_{ws}^1,T_{ws}^2,\cdots,T_{ws}^i,\cdots\}$ that is closest to the optimal threshold. The problem is how to enforce the monotonically decreasing of the threshold and how to stop the repeating procedure at the optimal point.
 
First, we initialize $T^1_{ws} = R\% \times N$, where $R$ is a user-specific hyper-parameter, e.g. $R=30$, and $N$ is the total number of snippets in the training dataset. This setting makes $T^1_{ws}$ a large enough threshold, since generally the ratio of abnormal data in the training dataset is less than 30\%. We test different $R$ (e.g., 35\%) in the experimental section and find that it does not influence our final results much, demonstrating the robustness of our method to the user interaction. 

Then, to ensure Eq.~\ref{eq:threshold_requirement}, our adaptive thresholding method determines the threshold used in the current module based on wOCC models trained in all previous modules. Each training module produces multiple wOCC models, obtaining one after each loop between wOCC and WS. 
Let $M_i$ be the number of wOCC models trained from module 1 to module $i$. When computing $T_{ws}^{i+1}$ for module $i+1$ with $i\geq 1$, we take all the $M_i$ wOCC models into consideration. Specifically, for each wOCC, e.g., the $j^{th}$ one where $j\in[1,M_i]$, we use it to identify $R\%$ of snippets in the training dataset that have high anomaly scores, and denote the resulting set of snippets as $A_j$. We then compute intersection between all sets and count the number of the resulting intersection as the threshold:
\begin{equation}
  T^{i+1}_{ws} = {\rm Num}(A_1\cap A_2\cap \cdots\cap A_{M_i}),
  \label{eq:G-i}
\end{equation}
where ${\rm Num}(\cdot)$ counts the number of elements in the intersection. The operation of computing intersection ensures that the obtained threshold is monotonically decreased. In essence, the threshold computed in this way means the number of snippets viewed as abnormal by all wOCC models. At the early stages, the number of wOCC models is small, therefore the size of the intersection is large, yielding large thresholds. As more wOCC models involved, it is more difficult to form a consensus, thus producing smaller thresholds. 

\begin{figure}[t]
    \centering
    \includegraphics[width=0.7\textwidth,height=4.5cm]{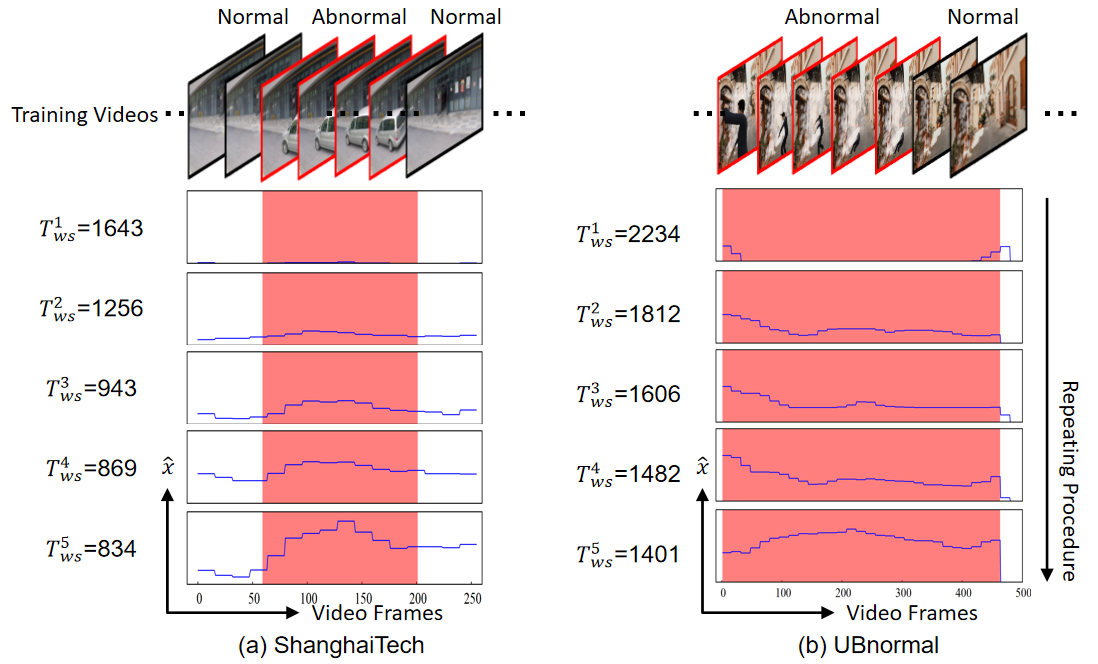}
    \caption{As repeating procedure goes, $T_{ws}$ monotonically decreases, improving the WS model gradually. Left: example from ShanghaiTech. Right: example from UBnormal.}
    \label{fig:vis_of_wx}
\end{figure}

To elucidate the effect of the adaptive thresholding, we visualize how WS models are gradually improved as the repeating procedure runs in Figure~\ref{fig:vis_of_wx}. In the figure, we show two examples. Taking the one on the left as an example, it shows a video with an abnormal event in the middle. At the beginning when $T_{ws}=1643$, the WS model wrongly identifies the abnormal event as normal. As the repeating procedure runs, $T_{ws}$ becomes smaller and smaller, which is finally reduced to 834 after repeating 5 times, and the WS model begins to identify the anomaly event progressively. 

\textbf{Stopping Criterion.}
As our goal is to find the threshold $T_{ws}^*$ close to the optimal threshold, we need to stop the repeating procedure at the right time. Experimentally, we find that $T_{ws}$ drops fast at the first few interleaving training modules, then the rate of change of $T_{ws}$ becomes smaller which indicates that the wOCCs achieve a consensus about anomalies. The size of the consensus set is probably the actual number of abnormal snippets in the dataset.  
Based on this observation, once change rate of $T_{ws}$ between two consecutive training modules is less than $Q\%$ of the first change rate at the very beginning, we stop the repeating procedure. Please see the high correlation between the change rate of $T_{ws}$ and the accuracy of the WS model in Figure~\ref{fig:convergence of G}.

\textbf{Training Time Analysis.} 
Though we need to repeat the interleaving training module multiple times, our training is as fast as the original wOCC or WS method. First, the repeating procedure usually stops with fewer than 6 modules thanks to the stopping criterion. Furthermore, we train wOCC (or WS) for just one epoch, and then exchange to train WS (or wOCC) for another epoch. If we have 6 modules, each running an average of 5 loops of interleaving training, the wOCC and WS models are trained for just 30 epochs each. The original wOCC or WS models are also trained for the similar number of epochs in their default experimental settings.  
The detailed experiments are in Section~\ref{sec:discussion}.

\subsection{Inference}
 At the inference stage, both wOCC and WS models can be used to detect anomalies. We provide results of both models when compared with previous methods. 

\section{Experiments}
\label{sec:experiment}
\subsection{Datasets and Evaluation Metrics}
\textbf{ShanghaiTech.}The ShanghaiTech dataset~\cite{liu2018future} contains 437 videos and was created as a benchmark for OCC with only normal videos available in the training set. Zhong et al.~\cite{zhong2019graph} reorganized the dataset to enable the training of WS systems. The new split contains 63 abnormal videos and 175 normal videos, while there are 44 anomalous and 155 normal videos in the new testing split. In our unsupervised setting, we follow the split for WS approaches but do not provide video-level annotations in the training stage.

\textbf{UBnormal.}The UBnormal dataset~\cite{Acsintoae_CVPR_2022} is a synthetic open-set benchmark, containing 268 training videos, 64 validation videos and 211 test videos. Normal and abnormal videos are mixed in the three subsets. The dataset is challenging because of disjoint sets of anomaly types in training and testing sets. We follow its original organization but train our proposed UVAD without any annotations.

\textbf{Evaluation Metrics.} Following prior VAD arts~\cite{tian2021weakly,hirschorn2023normalizing}, we use the frame-level area under ROC curve (AUC) to measure VAD's accuracy for both datasets.

\subsection{wOCC and WS models Employed}
As stated, our method is a flexible UVAD framework that can use different wOCC and WS models.  We test improving three OCC models to wOCC, including the AE model used in~\cite{zaheer2022generative}, Jigsaw~\cite{wang2022video}, and STG-NF~\cite{hirschorn2023normalizing}. \textit{The wOCC losses for AE and Jigsaw are put into supplemental material.}

(1) \textbf{AE.} GCL~\cite{zaheer2022generative} proposes to use an AutoEncoder that reconstructs the features extracted from videos as their OCC model. We use it here too.

(2) \textbf{Jigsaw.} The OCC model Jigsaw~\cite{wang2022video} addresses VAD by solving a pretext task: spatio-temporal jigsaw puzzles. It divides the spatio-temporal space of a video into smaller cubes, then shuffles the positions of the cubes, and finally tries to restore the original positions of the cubes.

(3) \textbf{STG-NF.} STG-NF~\cite{hirschorn2023normalizing} extracts people's pose sequences, and
extends Glow~\cite{kingma2018glow} to build multivariate Gaussian distribution of normal action sequences.

For WS, we test the method proposed in Sultani et al.~\cite{sultani2018real} and RTFM~\cite{tian2021weakly}. 

(1) \textbf{Sultani et al.}~\cite{sultani2018real}. Sultani et al.~\cite{sultani2018real} propose the first MIL VAD model that maximizes the separability between a positive bag containing snippets of an abnormal video and a negative bag with snippets of a normal video.

(2) \textbf{RTFM.} RTFM~\cite{tian2021weakly} extends Sultani et al.~\cite{sultani2018real}
to compare the Top-$k$ largest-magnitude snippets between positive and negative bags.

\subsection{Implementation Details}
We implement our method in PyTorch. For STG-NF and RTFM (implementation details of other wOCC/WS models are put in supplemental material), they are optimized by Adam optimizer with $\beta_1 = 0.9$, $\beta_2 = 0.999$, and a weight decay of 0.0005. The batch size of STG-NF is 256, and that of RTFM is 32. The learning rate is $5e-4$ and $1e-3$ for STG-NF and RTFM respectively. For RTFM, the margin $m$ is set to 100. During interleaving training, we train a model for one epoch and then exchange to train the other model for another epoch.

Our method has two hyperparameters shared by all types of wOCC or WS models. One is $R\%$ which determines the initial threshold $T^1_{ws}$. By default, we set it to 15\% for the ShanghaiTech dataset which contains a small ratio of abnormal data. The UBnormal dataset contains relatively more abnormal data and we set $R\%$ to $30\%$. The other parameter is $Q\%$ used to stop the repeating procedure. For both datasets, we set $Q\%$ to 10\%. Analysis of Q\% is put into \textit{Supp.}
\begin{table}[t]
\caption{Comparison with previous approaches. For wOCC, we use STG-NF~\cite{hirschorn2023normalizing}. For WS, we use RTFM~\cite{tian2021weakly}. We test using C3D~\cite{tran2015learning} and I3D~\cite{kay2017kinetics} to extract features for RTFM. Our$_{wOCC}$ gives the AUC of the wOCC model. Our$_{WS}$ gives the AUC of the WS model. The methods split by the dashed line are trained as a whole in our framework. * means we reimplement the method by taking I3D as the feature. ** means our method degenerated to the supervised OCC or WS method. *** means the results are reproduced by us using the official code.}
\raisebox{-1pt}{
    \begin{subtable}[c]{0.33\linewidth}
        \centering
        \scalebox{0.6}{
        \setlength{\tabcolsep}{0.1mm}{
        \begin{tabular}{l| c| c| c}
            \hline
            \multicolumn{4}{c}{\textbf{Unsupervised}} \\
            \hline
                 \textbf{Method} & \textbf{Features} & \textbf{\makecell{STech\\ AUC \%}} & \textbf{\makecell{UB \\ AUC \%}}\\
            \hline
       \makecell[l]{\rule[-7pt]{0pt}{20pt}AE$_{AllData}$} & I3D & 60.51 & - \\ \hline
        \makecell[l]{\rule[-7pt]{0pt}{20pt}STG-NF$_{AllData}$~\cite{hirschorn2023normalizing}} & - & 80.29 & 70.48 \\ \hline
        \makecell[l]{\rule[-7pt]{0pt}{19pt}GCL$^{*}$~\cite{zaheer2022generative}} & I3D & 76.14 & - \\ \hline
        \makecell[l]{\rule[-7pt]{0pt}{19pt}Our$_{wOCC}$} & - & 81.50 & 71.67\\ \cdashline{1-4}
        \makecell[l]{\rule[-7pt]{0pt}{19pt}Our$_{WS}$} & C3D & 85.43 & 59.05 \\ \hline
        \makecell[l]{\rule[-7pt]{0pt}{19pt}Our$_{wOCC}$} & - & 82.57 & \textbf{74.76}\\ \cdashline{1-4}
        \makecell[l]{\rule[-7pt]{0pt}{19pt}Our$_{WS}$} & I3D & \textbf{88.18} & 63.10 \\ \hline
        \end{tabular}
        }
        }
    \end{subtable}
        }
    \begin{subtable}{0.28\textwidth}
        \centering
        \scalebox{0.6}{
        \setlength{\tabcolsep}{0.1mm}{
        \begin{tabular}{l| c| c}
            \hline
                \multicolumn{3}{c}{\textbf{One-Class Classification}} \\
            \hline
             \textbf{Method} & \textbf{\makecell{STech\\ AUC \%}} & \textbf{\makecell{UB \\ AUC \%}}\\
            \hline
            MemAE~\cite{gong2019memorizing} & 71.20 & - \\ \hline 
            Frame Prediction~\cite{liu2018future} & 73.40 & - \\ \hline
            Markovitz et al.~\cite{markovitz2020graph} & 76.10 & 52.00 \\ \hline
            HF$^2$-VAD~\cite{liu2021hybrid} & 76.20 & - \\ \hline
            CT-D2GAN~\cite{feng2021convolutional} & 77.70 & - \\ \hline
            CAC~\cite{wang2020cluster} & 79.30 & - \\ \hline
            SSMTL~\cite{georgescu2021anomaly} &  82.40 & -\\ \hline
            Georgescu et al.~\cite{georgescu2021background} & 82.70 & 59.30\\ \hline
            SSMTL++~\cite{barbalau2023ssmtl++} &  83.80 & 62.10\\ \hline
            Jigsaw~\cite{wang2022video} & 84.30 & 56.40\\ \hline
            STG-NF~\cite{hirschorn2023normalizing} & 85.90 & 71.80\\  \hline
            Our$_{wOCC}^{**}$ (STG-NF) & \textbf{86.37} & \textbf{72.81} \\ 
            \hline
        \end{tabular}
                }
        }
    \end{subtable}
    \begin{subtable}{0.33\textwidth}
        \centering
        \scalebox{0.6}{
        \setlength{\tabcolsep}{0.1mm}{
        \begin{tabular}{l| c| c| c}
            \hline
            \multicolumn{4}{c}{\textbf{Weakly Supervised}} \\
            \hline
                 \textbf{Method} & \textbf{Features} & \textbf{\makecell{STech\\ AUC \%}} & \textbf{\makecell{UB \\ AUC \%}}\\
            \hline
            GCN~\cite{zhong2019graph} & C3D & 76.44 & - \\ \hline
            GCN~\cite{zhong2019graph} & TSN & 84.44 & - \\ \hline
            Zhang et al.~\cite{zhang2019temporal} & C3D & 82.50 & - \\ \hline
            Sultani et al.$^{***}$~\cite{sultani2018real} & I3D & 84.53 & 54.12 \\ \hline
            AR-Net~\cite{wan2020weakly} & I3D  & 85.38 & -  \\ \hline
            CLAWS~\cite{zaheer2020claws} & C3D & 89.67 & - \\ \hline
            MIST~\cite{feng2021mist} & I3D & 94.83 & - \\ \hline
            Li et al.~\cite{li2022self} & I3D & 96.08 & - \\ \hline
            RTFM$^{***}$~\cite{tian2021weakly} & C3D & 91.51 & 62.30\\ \hline
            RTFM$^{***}$~\cite{tian2021weakly} & I3D & 96.10 & 66.83\\ \hline
            S3R~\cite{wu2022self} & I3D & \textbf{97.48} & - \\ \hline
            Our$_{WS}^{**}$ (RTFM) & I3D & 96.33 & \textbf{67.42} \\
            \hline
        \end{tabular}
          }
        }
    \end{subtable}
    \label{tab:ShanghaiTech}
\end{table}
\begin{table}[!t]
    \begin{minipage}[c]{0.5\linewidth}
    \caption{Ablation study on wOCC improved from different OCC models, always with RTFM as the WS model.}
    \centering
    \scalebox{0.73}{
    \setlength{\tabcolsep}{0.5mm}{
    \begin{tabular}{c|cc|c}
    \hline
    OCC Model & Our$_{wOCC}$ & Our$_{WS} (RTFM)$ & GCL$_{Classifier}$ \\ \hline
    AE        & 70.99     & 78.90  & 76.14            \\ 
    Jigsaw~\cite{wang2022video}    & 81.23     & 85.35     & -                \\ 
    STG-NF~\cite{hirschorn2023normalizing}     & 82.57     & 88.18  & -                \\ \hline
    \end{tabular}
    }
    }
    \label{table:different-occ-model}
    \end{minipage}
    \begin{minipage}[c]{0.48\linewidth}
        \caption{Ablation study on using different WS models in our method, always with STG-NF as the wOCC model.}
        \centering
        \scalebox{0.76}{
        \setlength{\tabcolsep}{0.8mm}{
        \begin{tabular}{c|cc}
        \hline
        WS Model & Our$_{wOCC}$ (STG-NF) & Our$_{WS}$ \\ \hline
        \makecell{\rule[-5pt]{0pt}{16pt}Sultani et al. ~\cite{sultani2018real}}       & 81.92     & 77.41             \\ 
        \makecell{\rule[-5pt]{0pt}{16pt}RTFM~\cite{tian2021weakly}}     & 82.57     & 88.18                 \\ \hline
        \end{tabular}
        }
        }
        \label{table:different-ws-model}
    \end{minipage}
\end{table}
\subsection{Comparison with Previous Approaches}

From left to right, Table~\ref{tab:ShanghaiTech} shows the comparison between our and previous approaches under Unsupervised, One-Class Classification and Weakly-Supervised. Since the unsupervised methods are scarce, besides GCL~\cite{zaheer2022generative}, we compare with AE$_{AllData}$ which means training the AE model on the whole training dataset containing both normal and abnormal data. We also compare with STG-NF$_{AllData}$. For our method, we adopt STG-NF~\cite{hirschorn2023normalizing} as wOCC, and RTFM~\cite{tian2021weakly} as WS. 

First, Our$_{wOCC}$ (trained along with RTFM (I3D)) outperforms STG-NF$_{AllData}$ (see the table on the left). This validates the benefit of using our method to distinguish the normal from abnormal data in the training dataset. On ShanghaiTech, our proposed wOCC improves STG-NF from 80.29\% to 82.57\%. In contrast, the improvement is from 70.48\% to 74.76\% on UBnormal. The improvement on UBnormal is larger as there is more abnormal data in UBnormal which causes STG-NF$_{AllData}$ to perform worse. 

Second, please compare our method with GCL. Our$_{WS}$ achieves an AUC of 88.18\%, while that of GCL is 76.14\% (this score is reported by the classifier of GCL, thus we use our WS model for the comparison). This significant improvement is partly due to the reason that we use a stronger OCC model, \ie, STG-NF, than GCL. If using the same OCC model, our method outperforms GCL by 2.76\%, as shown in Table~\ref{table:different-occ-model}. 

Our UVAD method can be degenerated into a supervised OCC method. For example, we can perform our interleaving training on the normal data only. The obtained wOCC model in this setting is denoted by Our$_{wOCC}^{**}$ which outperforms existing supervised OCC approaches (see the table in the middle). This validates the effectiveness of the proposed wOCC with weighted importance, as even among the normal data there is data that is more normal than other data, and the wOCC model can better identify their difference.

To degenerate our method into a supervised WS method, we allow our WS model to know video-level labels. Specifically, snippets of normal videos are treated as normal. Since snippets of abnormal videos can be either normal or abnormal, they are processed by our adaptive thresholding approach.  
As seen in Table~\ref{tab:ShanghaiTech} on the right, Our$_{WS}^{**}$ (RTFM) outperforms the baseline RTFM~\cite{tian2021weakly}. On UBnormal, our method performs best among all the compared WS methods.

\subsection{Ablation Study and Discussions}
\label{sec:discussion}
Next, we perform ablation studies on our method. Without extra explanations, experiments are conducted by taking STG-NF as wOCC and RTFM as WS.  

\begin{figure}[t]
    \begin{minipage}[c]{0.49\linewidth}
        \centering
        \includegraphics[height=3.3cm]{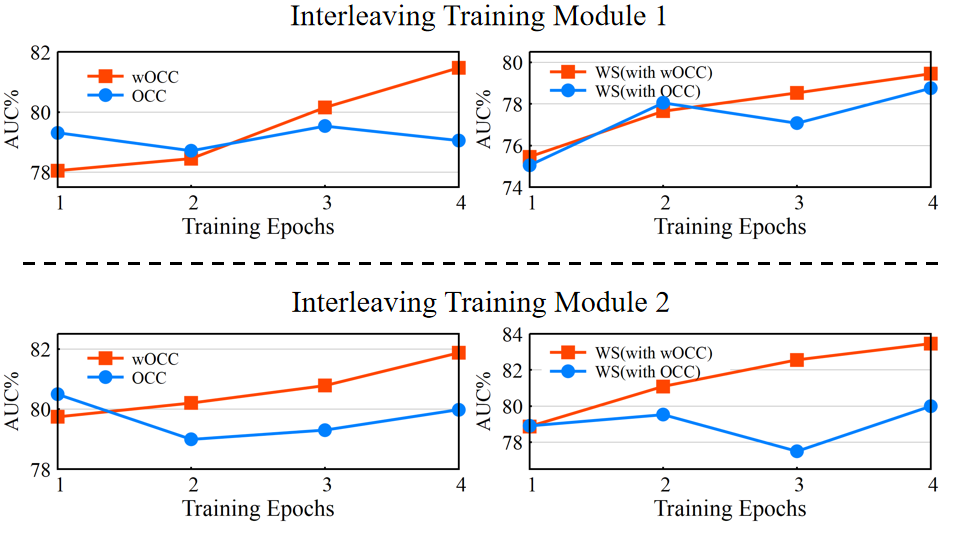}
        \caption{Performance fluctuation in the first and second modules. Red line: AUC of wOCC (or WS) when interleaving wOCC and WS. Blue line: AUC of OCC (or WS) when interleaving OCC and WS.}
        \label{fig:stable}
    \end{minipage}
    \begin{minipage}[c]{0.49\linewidth}
        \centering
        \includegraphics[width=0.49\columnwidth,height=3.3cm]{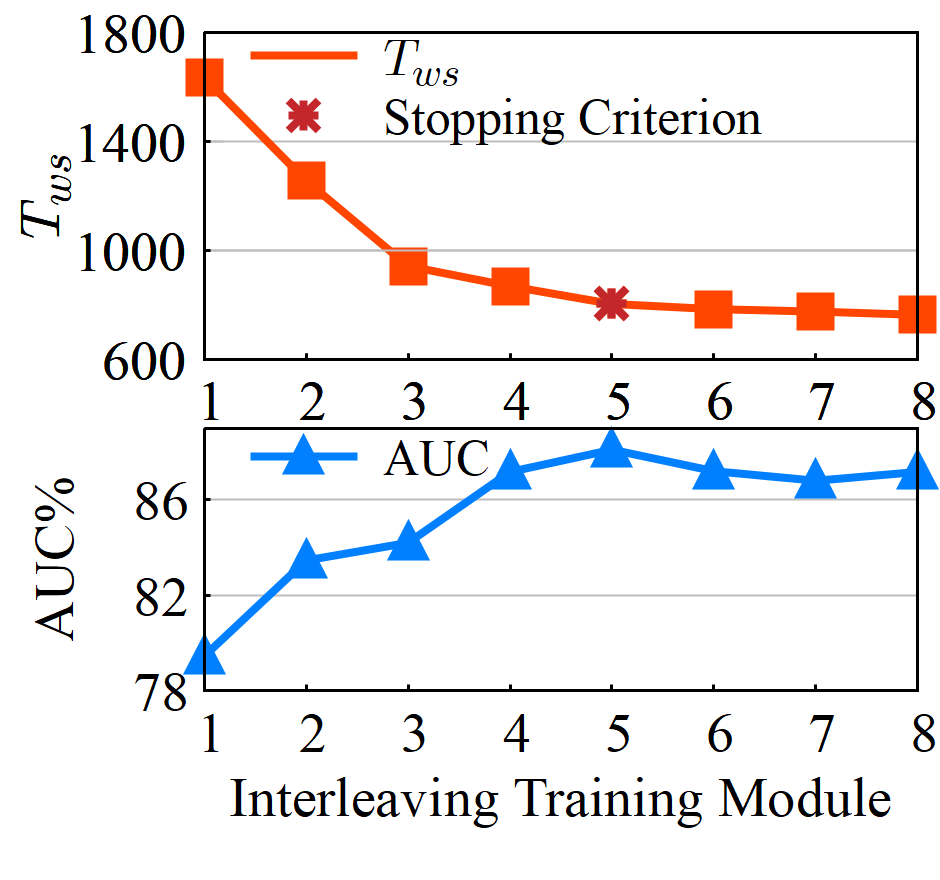}
        \includegraphics[width=0.49\columnwidth,height=3.3cm]{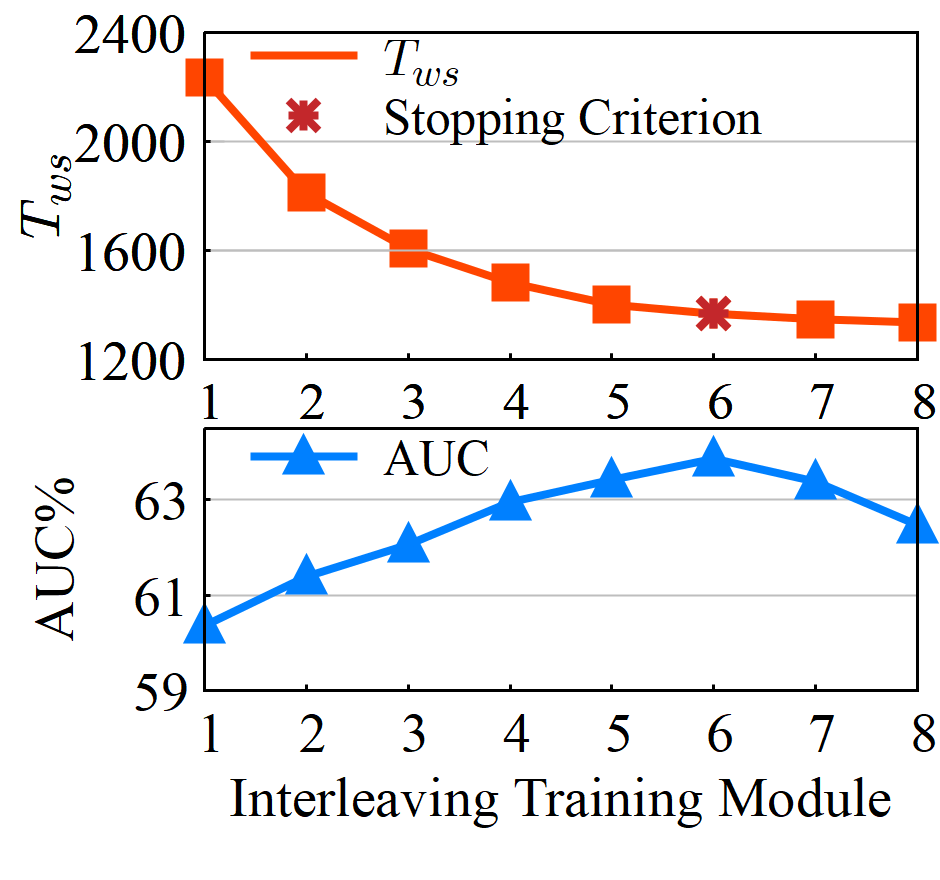}
        \caption{Left: ShanghaiTech. Right: UBnormal. Top: $T_{ws}$ in each interleaving training module. Bottom: The AUC of the WS model in each module. Stars in top figures: position where achieves stopping criterion.}
        \label{fig:convergence of G}
    \end{minipage}
\end{figure}

\textbf{Comparison between OCC over wOCC.} In Figure~\ref{fig:stable}, we compare the performance fluctuation of using wOCC or OCC (STG-NF) in our framework in the first two interleaving training modules' training. As shown, directly using OCC leads to performance fluctuation of both OCC and WS models and converges to an unsatisfactory point. In comparison, our proposed wOCC stabilizes the training and improves the performance of both models.

\textbf{Using Different OCC/WS Models.} In Table~\ref{table:different-occ-model}, we conduct experiments of upgrading different types of OCC models to wOCC in our method. We test AE~\cite{zaheer2022generative}, Jigsaw~\cite{wang2022video}, and STG-NF~\cite{hirschorn2023normalizing}, while always using RTFM~\cite{tian2021weakly} as the WS model. As seen, the effectiveness of our method is essentially affected by wOCC model. Similarly in Table~\ref{table:different-ws-model}, the same wOCC model is used but with different WS models. Overall, a better wOCC or WS model yields better results. 
\begin{table}[t]
    \caption{Comparison with fixed thresholding. 
    }
    \centering
    \scalebox{0.8}{
    \setlength{\tabcolsep}{3mm}{
    \begin{tabular}{c c c c c}
    \hline
    Weighted OCC & WS with Adaptive Threshold & RTFM (AUC \%) & STG-NF (AUC \%)\\ \hline
     \ding{55}& \ding{55} & 82.06 & 80.52\\
     \checkmark & \ding{55} & 83.48 & 81.78 \\
     \ding{55} & \checkmark & 85.86 & 81.94 \\
     \checkmark & \checkmark & \textbf{88.18} & \textbf{82.57} \\
     \hline
    \end{tabular}
    }
    }
    \label{tab:ablation-on-fixed-thresholding}
\end{table}

\textbf{Comparison with Fixed Thresholding.} In Table~\ref{tab:ablation-on-fixed-thresholding}, the first row shows results of using fixed thresholding for both OCC and WS models. The second and third rows use fixed thresholding for either OCC or WS models. The fourth row shows our method with wOCC and adaptive thresholding. As seen, our method outperforms all the other variants.

\textbf{Effectiveness of the Stopping Criterion.} In Figure~\ref{fig:convergence of G}, we show that with the current stopping criterion, our method can stop when it achieves the best AUC (\ie, the best VAD accuracy). As the repeating procedure runs, $T_{ws}$ drops and AUC increases. $T_{ws}$ drops very fast at the very beginning, and then the drop rate becomes slower. Simultaneously, AUC first increases and then decreases. We stop our method when the change rate of $T_{ws}$ is about 10\% of its original change rate, which finds the best stopping positions indicated by the stars in top figures.

\begin{figure}[t]
\begin{minipage}[c]{0.48\linewidth}
        \centering
        \includegraphics[width=0.6\textwidth,height=2cm]{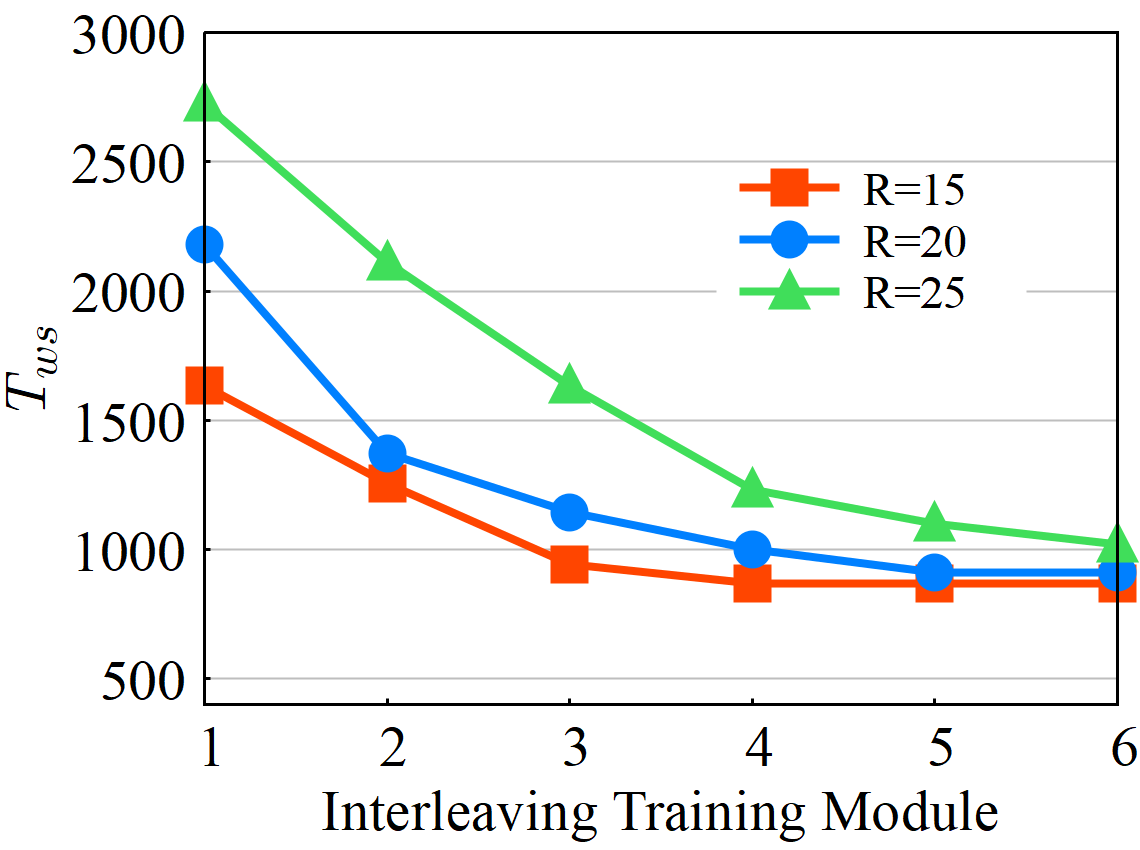}\\
        \includegraphics[width=0.49\columnwidth,height=2cm]{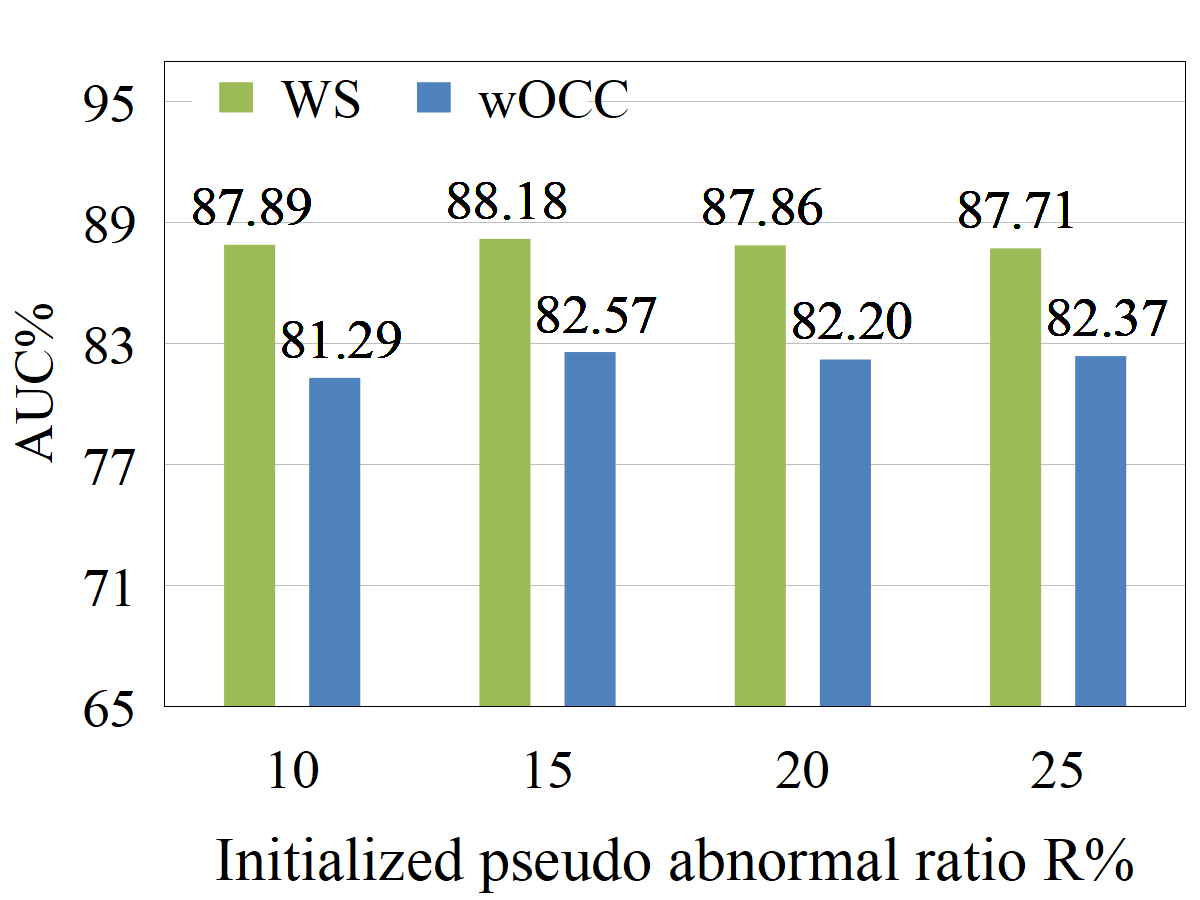}
        \includegraphics[width=0.49\columnwidth,height=2cm]{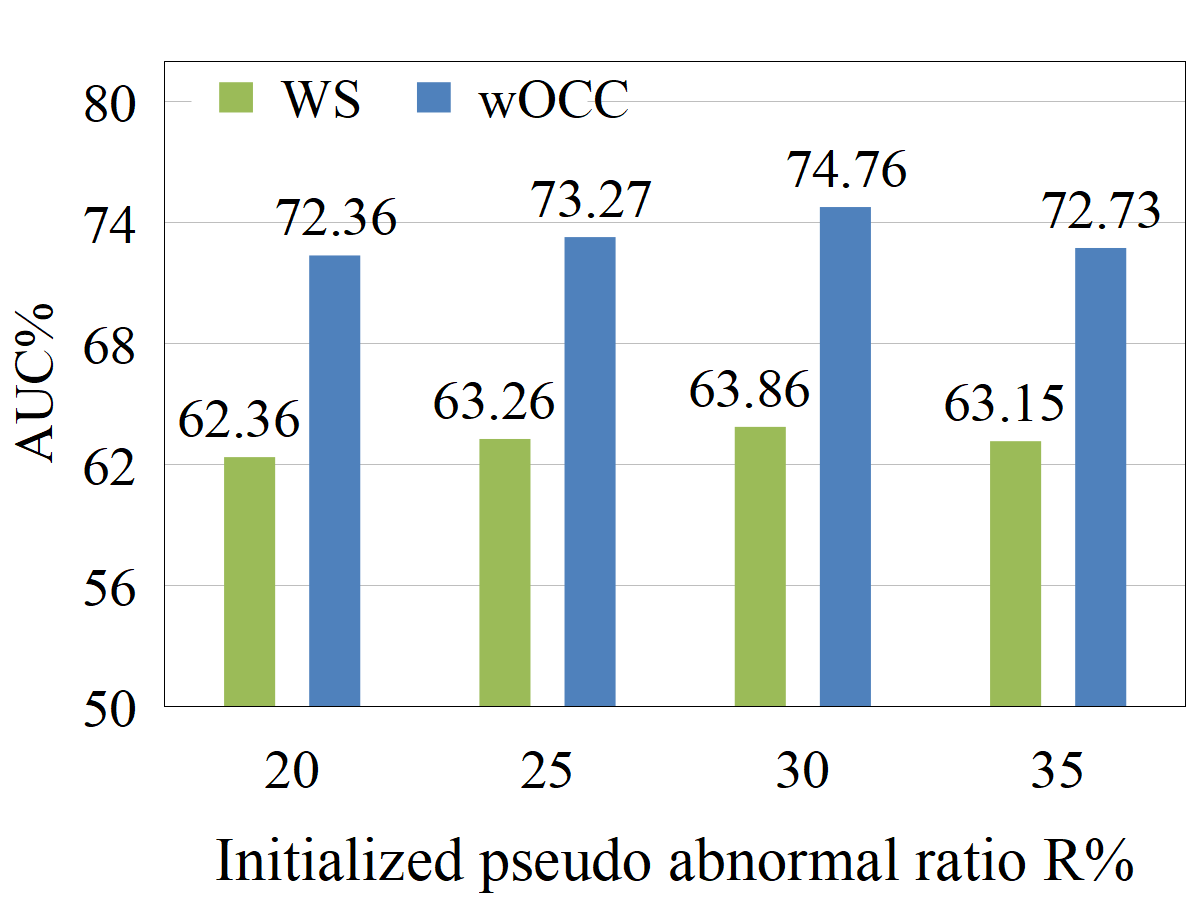}
        \caption{Top: different $R$\% converge to similar $T_{ws}$ (on ShanghaiTech). Bottom: different $R$\% yield similar AUC (Left: on ShanghaiTech, Right: on UBnormal).}
        \label{fig:convergency-of-R}
    \end{minipage}
    \begin{minipage}[c]{0.48\linewidth}
        \centering
        \includegraphics[width=1\columnwidth,height=4cm]{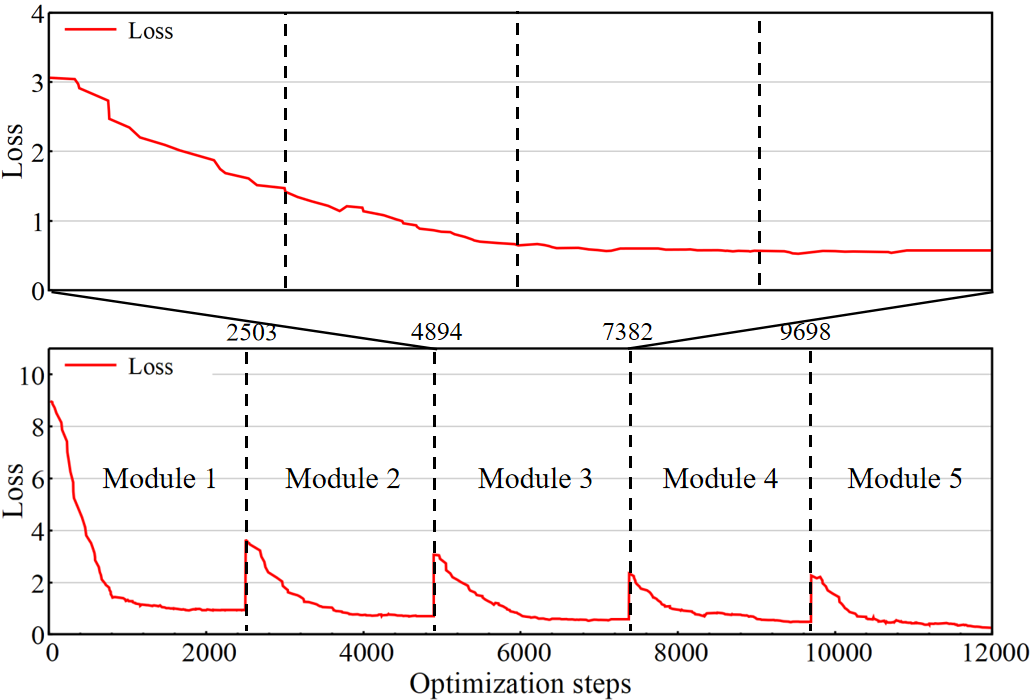}
        \caption{Bottom: training loss curve of the wOCC model during the whole repeating procedure. Top: zoomed-in training loss curve in the third module.}
        \label{fig:training-loss-curve}
    \end{minipage}
\end{figure}

\textbf{Robustness to $R$\%.} Our method is robust to the parameter $R$\%, as demonstrated in Figure~\ref{fig:convergency-of-R}. On the top, we show that the threshold $T_{ws}$ converges to similar values after 6 interleaving training modules when using different $R$\%. At the bottom, we show that different $R$\% yield similar AUC on both training datasets. The experiments demonstrate that our method is not influenced much by this parameter for each dataset. For different datasets, we need to set different $R$\% that roughly match with the ratio of anomaly data in the datasets.

\begin{table}[t]
    \caption{Training time on ShanghaiTech.}
    \centering
    \scalebox{0.7}{
    \setlength{\tabcolsep}{3mm}{
    \begin{tabular}{c|c|c|c|c|c|c}
    \hline
    Model & \makecell{Epoch \\ Each Step} & \makecell{Training\\Steps} & \makecell{Training\\Epochs} & \makecell{Training\\ Time}& \makecell{STG-NF\\AUC \%} & \makecell{RTFM\\AUC \%} \\ \hline
    \multirow{3}{*}{Our UVAD} & 1 & 17 & 2$\times$1$\times$17 & 2h28m & \textbf{82.57} & \textbf{88.18} \\ \cline{2-7}
    & 2 & 12 & 2$\times$2$\times$12 & 2h42m & 82.53 & 87.95 \\ \cline{2-7}
    & 3 & 11 & 2$\times$3$\times$11 & 3h31m & 82.27 & 86.36 \\ \hline
    STG-NF~\cite{hirschorn2023normalizing}& - & - & 8 & 10m & 85.90 & -\\ \hline
    RTFM~\cite{tian2021weakly}& - & - & 50 & 2h32m & - & 97.21 \\ \hline
    \end{tabular}
    }
    }
    \label{tab:training-time}
\end{table}

\textbf{Training Loss Curve.} In Figure~\ref{fig:training-loss-curve}, we show training loss curves of the wOCC (STG-NF) model in our framework. As seen, the loss drops smoothly within each interleaving training module. Since we re-train the wOCC model from scratch in the next module, the loss suddenly increases at the beginning of the next module, but the peak magnitude of the loss is lower than that of the previous module. Please check the zoomed-in curve showing the loss curve of the wOCC model in the third module. Even though the model is trained alternately with another model, its loss can still decrease smoothly, not affected by the other model. The WS model's training loss curve is put into the supplemental material.

\textbf{Training Time.} As mentioned in Section~\ref{sec:adaptive_thresholding}, although we need to train the wOCC and WS models alternately many times in multiple modules, our method is fast in training. That is because, we just train a model for one epoch and then exchange to train the other model for another epoch, which form a training step of our method. Our interleaving training converges very fast. As shown in Table~\ref{tab:training-time}, there are only 17 training steps during the whole training, yielding a total of 34=2(wOCC and WS)$\times$1(epoch)$\times$17(all 
training steps) epochs. It takes around 2.5 hours to train our method on ShanghaiTech. 
We have also trained the wOCC or WS model for 2 or 3 epochs in each training step, but obtained worse results, probably because training with more epochs would enhance the supervision of wrong pseudo labels in early training modules.

\section{Conclusion}
\label{sec:conclusion}
In this paper, we interleave the One-Classification and Weakly-Supervised models with adaptive thresholding to tackle the unsupervised video anomaly detection problem without human annotation. We face the challenges of performance fluctuation and the demanding of accurate threshold when designing the framework. To alleviate performance fluctuation, We propose the wOCC model which requires soft labels but not binary labels. To obtain a relative optimal threshold, we repeat our proposed interleaving training module, during which we propose a monotonically threshold decreasing mechanism to refine a rough threshold progressively. Extensive experiments demonstrate the effectiveness of our method. Remarkably, our method can be upgraded with the most recent development in OCC and WS VAD fields.

\section*{Acknowledgements}

This work was partly supported by the Natural Science Foundation of China (62072191, U21A20520, 62325204), the National Key Research and Development Program of China (2022YFE0112200), the Key-Area Research and Development Program of Guangzhou City (202206030009), and the Guangdong Basic and Applied Basic Research Fund (2023A1515030002, 2024A1515011995).
%
%
\bibliographystyle{splncs04}
\bibliography{main}

\end{document}